# Characterizing the Emotion Carriers of COVID-19 Misinformation and Their Impact on Vaccination Outcomes in India and the United States

Ridam Pal[1*], Sanjana S[2*], Deepak Mahto[1], Kriti Agrawal[3], Gopal Mengi[1], Sargun Nagpal[1], Akshaya Devadiga[1], Tavpritesh Sethi[1#]

[1]Indraprastha Institute of Information Technology, Delhi, India
[2] PES University, Bangalore, India
[3] BITS Pilani, K.K. Birla Goa Campus, Goa, India
* Equal Contribution

**Corresponding Author:** Dr. Tavpritesh Sethi, Associate Professor, IIIT Delhi, New Delhi, 110020
[#]**Corresponding Author's Email:** tavpriteshsethi@iiitd.ac.in

## Abstract

**Background:** The COVID-19 Infodemic had an unprecedented impact on health behaviors and outcomes at a global scale. While many studies have focused on a qualitative and quantitative understanding of misinformation, including sentiment analysis, there is a gap in understanding the emotion-carriers of misinformation and their differences across geographies. In this study, we characterized emotion carriers and their impact on vaccination rates in India and the United States. Our large language model was trained on 2.3 million tweets carrying COVID-19 misinformation and validated prospectively on a manually annotated set of 2000 misinformation tweets across the globe.

**Methods:** A manually labelled dataset was created and collated with three publicly available datasets (CoAID, AntiVax, CMU) to train deep learning models for misinformation classification. Misinformation labelled tweets were further analyzed for behavioral aspects by leveraging Plutchik Transformers to determine the emotion for each tweet. Time series analysis was conducted to study the impact of misinformation on spatial and temporal characteristics. Further, categorical classification was performed using transformer models to assign categories (Authority, Symptoms, Cure, Vaccination and Spread) for the misinformation tweets.

**Findings:** Word2Vec+BiLSTM was the best model for misinformation classification, with an F1-score of 0.92. The US had the highest proportion of misinformation tweets (58.02%), followed by the UK (10.38%) and India (7.33%). Disgust, anticipation, and anger were associated with an increased prevalence of misinformation tweets. Disgust was the predominant emotion associated with misinformation tweets in the US, while anticipation was the predominant emotion in India. For India, the misinformation rate exhibited a lead relationship with vaccination, while in the US it lagged behind vaccination.

**Interpretation:** Our study deciphered that emotions acted as differential carriers of misinformation across geography and time. These carriers can be monitored to develop strategic interventions for countering misinformation, leading to improved public health.

**Funding:** No funding was received.

# Research in context

- **Evidence before this study**

    Numerous studies have brought to light the substantial role that "misinformation" plays in influencing public perceptions, behaviour, and pandemic-related health outcomes. A PubMed search with the keywords [COVID-19] AND [Misinformation] yielded 17,171 results, indicating the significant interest and attention given to this subject. However, the number of pertinent studies dropped to 1,732 results when further filters, such as [Emotion] were added to the search criterion. It is essential to highlight that the majority of these studies were either reviews or focused on certain topics, such as mental health and effective communication. By adding location-specific filters, such as [India] and [USA], the search was further refined to produce 64 and 294 results respectively. These findings draw attention to the importance of studies particularly evaluating the relationship between COVID-19 misinformation, emotions, and vaccination outcomes.

- **Added value of this study**

    This study aimed to determine the severity of COVID-19 misinformation and its implications. It investigated the relationship between differential emotional carriers of misinformation with respect to temporal and spatial changes. Furthermore, we studied the impact of both misinformation rates and differential carriers of misinformation on vaccination rates. Additionally, the study conducted a temporal analysis of the categorization of misinformation tweets, investigating their dynamic evolution over time and their alterations in response to significant events. To our knowledge, this research is the first examination of emotional carriers of misinformation and their context with respect to vaccination rates in India and the United States. This end-to-end framework was built using state-of-the-art techniques to facilitate the identification of (mis)information, analyse its emotional content and depict to which category it belonged. The comprehensive framework can be extended to other diseases such as Tuberculosis, and mental health, thereby providing crucial support to healthcare professionals in promoting accurate information about the specific disease.

- **Implications of all the available evidence**

    In the current social media driven world, there is a prolonged risk of misinformation, severely affecting several aspects of the society. On probing, we found that there were very few studies conducted to comprehend misinformation and its association with human emotion and vaccination rates. To tackle the issue of misinformation, we require data-driven modeling approaches, which would curb the impact of misinformation on public health and help in developing Twitter guidelines to abate its spread.

# Introduction

With the outbreak of COVID-19, the World Health Organization emphasized the need for accurate and timely information dissemination for combatting the infodemic[1]. Social media played an important role in propagating information and shaping public perceptions[2]. Various surveys reported an increase in the usage of social media during the phase of national lockdown in several countries [3,4]. Social Media has been a significant contributor to the infodemic[5] and its role has been addressed by numerous studies [6–8]. However, the challenge to mitigate the infodemic extends far beyond eliminating fake news or rumors from social media. It encompasses a wide range of malicious content which includes conspiracy theories, panic, racism, xenophobia, and mistrust in the authorities, among others [9,10]. Additionally, the global infodemic led to undesired mental behavior and poor well-being [11–13]. Although studies have extensively focussed on detecting misinformation, there is a lack of understanding of the generic features of misinformation and their impact on public health outcomes.

Misinformation was identified as the primary contributor for spreading global infodemic[14]. In prior literature, misinformation has been defined based on a diverse range of perspectives and insights. In a general sense, it was defined as (i) "Objectively incorrect information, as determined by the best available evidence and expertise on the subject"[15]; (ii) "False information shared by people who do not intend to mislead others"[16]; (iii) "False and inaccurate information that is spread intentionally or unintentionally"[17]. In an effort to define health misinformation, certain authors proposed alternate definitions of misinformation from a healthcare perspective. They described health misinformation as (i) "A health-related claim of fact that is currently false due to a lack of scientific evidence"[18]; (ii) "Information that is contrary to the epistemic consensus of the scientific community regarding a phenomenon"[19]. Our study intends to identify tweets that can be classified as misinformation based on their relevance to healthcare. We defined misinformation as health-related information which is currently false or inaccurate due to lack of scientific evidence, spreading without any specific intent.

Contrary to this, NOT misinformation was defined as unchecked information in any given context. The rampant increase in social media engagement during the COVID-19 pandemic led to an alarming rise in misinformation[20]. Healthcare organizations, policymakers, and researchers soon recognized the urgent necessity to address this issue. Prior studies were conducted to investigate the grim effects of misinformation and its impact on public health and human behavior[6–8,21,22]. The conclusion of these studies depicted that misinformation resulted in the proliferation of negative emotions such as confusion, fear, panic, and anxiety and posed challenges for health officials to disseminate accurate information to the public. It reduced adherence to public health guidelines and mistrust of healthcare authorities. Additionally, misinformation was linked to unnecessary hospitalizations and overuse of medical resources, which immensely strained the healthcare system[23,24]. Several events have been reported in the study "The 'Pandemic' of Disinformation in COVID-19"[25], where mass media channels have shared incomplete or unverified updates on new treatments, myths about the use of masks, and errors by some hospital organizations that resulted in higher reluctance from patients to seek medical attention. These findings highlight the need for effective strategies to limit the dissemination of false or misleading information and mitigate the impact of misinformation on social media to ensure the well-being of individuals and the broader community.

To abate the spread of misinformation on social media, efforts have been made to classify them, employing deep learning, and transformer-based architectures [26,27]. However, the curation of misinformation datasets is a tedious task due to the need for benchmark datasets. P. Singh et al. outlined approaches for curating misinformation datasets specific to COVID-19 and how it can be re-created locally[28]. Extensive research efforts have been conducted to gain insights into the characteristics of misinformation tweets. These investigations have focused on various aspects, including the level of user engagement with misinformation, the cross-country dissemination patterns of misinformation, and the profound impact of misinformation on human behavior and vaccination rates [6,14,29–31]. Many studies were conducted on topic modeling and sentiment analysis on Twitter data across various geographic locations, which deciphered human behavior in response to specific events of the COVID-19 pandemic [32,33]. Also, multilingual approaches were adopted to intervene in the spread of misinformation tweets while understanding the sentiments associated with it and categorizing it into predominant topics [34–36]. Lately, W. Ying et al. presented an overview of the worldwide issues provided by the COVID-19 infodemic, using the psychological entropy model. The study suggested viable remedies such as the usage of technology-based measures to detect and control fake news on social media, utilization of risk communication to bridge the gap between awareness and uncertainties, and cultivation of eHealth literacy among the public[9].

While considerable studies have been conducted to mitigate misinformation related to COVID-19, several research gaps still need to be addressed. Although these studies analyze misinformation dissemination, they do not examine its impact on public health outcomes such as vaccinations and testing rates. Furthermore, while these studies perform binary classification to predict if a tweet contains misinformation, classifying misinformation into topics such as causes, spread, and vaccinations can aid in devising more informed intervention strategies to curb its spread. Additionally, the dynamics of topics related to COVID-19 change over time as new information emerges based on evolution in public health guidelines and shifts in societal attitudes. For instance, terms such as "AstraZeneca", and "Covaxin" had limited usage before the release of COVID-19 vaccines in India. Therefore, supervised machine learning methods trained on data from a limited time span are not robust to data-drift, and need to be periodically re-trained. Additionally, prior approaches for sentiment analysis have primarily focused on categorizing misinformation into broad sentiment categories, specifically positive, negative, or neutral[37]. As a result, they failed to capture fine-grained emotions associated with psychological factors such as disgust, fear and anxiety. The emotions expressed in misinformation tweets can be predictive of behavioral outcomes. Finally, most studies analyzed misinformation on limited data collected over a constrained time frame and geographical focus, prominently in countries like USA, UK and Brazil, potentially leading to outdated conclusions [6,32,36]. The results of such studies will not accurately reflect the impact of COVID-19 misinformation on a global landscape, essentially hindering policy-makers to design robust mitigation measures.

In this study, we analyzed a dataset of 2.3 million COVID-19 tweets carrying misinformation labels, spanning 168 countries and three COVID-19 waves (the first, omicron, and delta wave) to characterize and quantify emotion-carriers of misinformation and their impact on vaccination rates. We hand-labelled 2000 tweets and aggregated our annotated dataset with publicly available labeled datasets to study the global spatio-temporal spread of misinformation. The availability of a large Twitter dataset as this for retrospective analysis is key for deciphering the effect of misinformation on a global landscape. We examined fine-grained emotions in COVID-19 tweets and studied their role as carriers of misinformation. Additionally, we performed category classification using a robust keyword-similarity approach, facilitating seamless inclusion of new keywords to update topics as they evolve, based on domain-expertise or data-driven approaches. Finally, we studied the lead-lag relationship of misinformation with public health behavioral outcomes such as vaccinations, and caseloads in India and the United States. Our study provides valuable insights for developing targeted interventions to effectively abate the spread of misinformation on social media platforms and minimize its detrimental impact on public health.

## Methods

**Study Design and Data source**

In this study, we created a human labelled annotated dataset of 2000 tweets and aggregated it with three publicly available labeled datasets (i.e. CoAID[38], Antivax[39], and CMU[40]) to train deep learning models for misinformation classification. These labelled datasets collectively formed the discovery cohort for model training. The distribution of (mis)information tweets across the training set and testing set of the discovery cohort is provided in Supplementary Table 2 and Supplementary Table 3 respectively. We accumulated 2.3 million unlabelled COVID-19 related tweets from Panacea lab[41] and 16,520 tweets from the aforementioned publicly available datasets. Additionally, we retrieved tweet metadata (user id, user name, location, date, retweet count, favorite count, and user-verified information) using Twitter API (Tweepy). The attributes present in the cohort are illustrated in Supplementary Table 1. The inclusion criteria for the tweets were (i) availability of full text; (ii) identified as being in the English language; (iii) availability of accurate geo-location; (iv) did not violate COVID-19 Twitter policies.

We reserved a test set of 25% human labelled tweets (500 tweets) for prospective validation of our trained models. The inclusion of 1500 manually annotated tweets to the discovery cohort enabled diversification during training, effectively helping to build robust deep learning models. We partitioned the discovery cohort into an 80%-20% train-validation split for evaluating our models as shown in Supplementary Table 4. The best-trained model (fastText+BiLSTM) was used to label the 2.3 million tweets in Panacea lab (termed as a retrospective cohort) to perform retrospective spatio-temporal analysis of misinformation propagation. In addition, we trained deep learning models to perform emotion and topic classification on this dataset. Thereafter, we collected public health data on vaccinations and positive case rates from data.org [42], to understand the impact of misinformation and emotions as differential carriers of misinformation for the USA and India.

**Preprocessing**

The discovery and retrospective cohorts were cleaned to eliminate noise for extracting relevant information. The tweets were pre-processed using the regex library in Python to remove HTML characters, hyperlinks, punctuations, and mentions. Stop words were removed using the NLTK library so the models could focus on informative words relevant to the tasks. The tweet texts were converted to lowercase, and the words were reduced to their roots using the PortStemmer() function from the NLTK library[43].

**Procedure and Intervention**

**Misinformation Classification**

A compendium of 10 deep learning models was trained on the discovery cohort to classify misinformation. This work employed two approaches for training the models. The first approach involved the models being trained through transfer learning. For this, a pre-trained unsupervised embedding matrix for the vocabulary was generated from the language models namely, Word2Vec, fastText, GloVe, and BERT. These matrices were fed as input to the input layer of the recurrent neural network. Bi-directional LSTM was used as the deep learning architecture for these models, where the generated unsupervised embeddings were propagated through the BiLSTM layers followed by an output layer with softmax activation function for two-label classification. The second approach was training models based on transformer architectures like RoBERTa, SapBERT, SpanBERT, and Biosyn-SaBERT, which employed a self-attention mechanism for providing context-aware representation of tweet text. The BERT-based models were implemented using the HuggingFace library[44].

Misinformation classification for retrospective data was achieved through a human-in-the-loop process. 502 tweets were randomly sampled from the retrospective cohort and manually annotated by two annotators. The annotations that both annotators agreed upon were considered for testing. Cohen's Kappa score between the annotators was 0.96, indicating a near-perfect consensus. This data was used as a development set for the top 3 previously benchmarked models trained on the discovery cohort. The labels for the complete retrospective cohort (~2.3M tweets) were generated using the model which performed best on the randomly sampled tweets from the retrospective cohort (development set). Table 1 shows the instances of misinformation and not misinformation in coherence with our dataset.

| Class | Description | Tweets (Example) |
|---|---|---|
| 0 | Misinformation | UPI News: 18% of COVID-19 deaths in the U.S. linked to air pollution, study finds |
| 1 | Not Misinformation | BBC News - Coronavirus: Starbucks closes 2,000 Chinese branches |

**Table 1:** Description of the labels used for misinformation classification

**Emotion Classification**

To analyze the impact of misinformation on public opinion about the pandemic and its long-term effect on society, emotions for each tweet across both cohorts were generated. The emotions were classified into eight primary categories using Plutchik's wheel of emotion. The primary emotions included joy, trust, fear, surprise, sadness, anticipation, anger, and disgust. To calculate the emotion of each tweet, the Plutchik Transformer was used. This transformer model provided a fine-grained analysis of sentiments by correlating lexicons with specific emotions. The pre-trained weights of the Plutchik transformer were used to obtain a probability distribution over the eight emotions for the tweet text.

**Category Classification**

We utilized the IFCN dataset, majorly consisting of seven topics: Vaccine, Authorities, Conspiracy Theory, Cures, Spread, Causes, and Symptoms, to extract keywords relevant to these categories. To preprocess the data, we employed standard techniques such as NLTK stopword removal and digit removal using regular expressions. We aggregated all the text documents related to each topic, ensuring the creation of unique documents for each category. Next, we applied the CountVectorizer method to obtain a term-frequency matrix, considering both unigrams and bigrams. This matrix allowed us to quantify the frequency of occurrence of words and word pairs within the dataset. Based on this matrix, we employed a modified variant of the c-TF-IDF (class-based Term Frequency-Inverse Document Frequency) approach, to identify the most significant keywords for each category. After obtaining the top words using c-TF-IDF, we conducted manual validation and made necessary additions and elimination to ensure the relevance of the extracted keywords. Furthermore, the tweets were classified into categories based on the keywords extracted across each category to understand their related context to topics. For category classification, we developed three models, viz-a-viz, BERT, Bio Bert, and CT Bert. The following workflow was maintained to conduct category classification - (i) For individual models, embeddings for each word in the preprocessed tweet were calculated. (ii) After extracting embeddings for individual words in a tweet, tweet embedding was created by averaging all the individual word embeddings. (iii) In a similar process, embedding for each seed word in every category was calculated (iv) Cosine similarity between the tweet embedding and all the keyword embeddings was calculated. (v) Empirically, after experimenting, a threshold of 0.75 was set as a cut-off. (vi) The number of keywords in each topic that had a cosine similarity of greater than 0.75 with the tweet embedding was calculated. (vii) The category which had the maximum number of words crossing the cut-off threshold was selected as the category for the tweet. In case the number of words crossing the threshold was equal for more than one topic for a particular tweet, a topic from this list was randomly selected.

Based on the keywords, a corpus of 10 tweets per topic was created as a baseline for evaluating each model. Ten tweets per topic were manually selected, and each model's prediction was compared against our classification. As per our evaluation, Bert Base gave us the best results. Furthermore, based on the efficacy, the BERT base model was chosen to classify ~2.3 million tweets from the retrospective cohort. Table 2 shows some of the selected seed words from each category.

| Authorities | Symptoms | Cures | Vaccine | Spread |
|---|---|---|---|---|
| president | breath | ivermectin | vaccination | bat |
| government | cold | hydroxychloroquine | pfizer | infected |
| prime minister | runny nose | zinc | inovio | Wuhan market |
| state | lung | treatment | astrazeneca | transmitted |
| country | nose | medicine | moderna | face mask |

**Table 2:** Selected seed words with respect to each category for category classification

**Spatio-temporal analysis**

We examined the impact of misinformation by conducting spatial and temporal analysis on tweets across both the cohorts. For spatial analysis, tweets were stratified based on geographic location using the Open street map API Nominatim geocoder. While for temporal analysis, dates were extracted from each tweet for segregation. We analyzed the top five countries (United States, Canada, United Kingdom, India, and Australia) with the most misinformation tweets.

**Ethics Approval and Funding Source**

No ethics approval is required. No funding was received by any organization specifically for this analysis.

# Results

## Data characteristics and descriptive analysis of cohort

2,243,512 tweet IDs met the inclusion criteria from 353,350,148 IDs in the filtered Panacea lab dataset collected between February 2020 and October 2022. Based on the performance assessment of 500 randomly sampled tweets from the retrospective cohort, fastText+BiLSTM was chosen for labeling the complete (~2.3M) retrospective cohort. 243,157 (5.34%) tweets were labeled as misinformation, whereas 2,000,355 were labeled as not misinformation. While US citizens tweeted the most tweets overall (47.13%), the distribution of misinformation tweets was similar approximately across different countries as seen in Table 3. Predominantly, a high frequency (533,412) of COVID-19 tweets were tweeted during the Delta variant wave where the misinformation rate (7.04%) was considerably higher than other waves.

|  |  | **Misinformation (%)** | **(Not)misinformation (%)** | **Misinformation (Tweet)** | **(Not)misinformation (Tweet)** |
|---|---|---|---|---|---|
| **Total** | Total Tweets (22,43,512) | 5.3424 (1,19,858) | 94.6575 (21,23,654) | vaccine mandates have been a thing for over years to save lives and stop disease how'd we get here | what's hard to understand about no dine in just move it to take out lol |
| **Country** | India (1,84,516) | 4.3806 (8,083) | 95.6193 (1,76,433) | wishing early recovery | thats chinas total spike but yet we will shamelessly blame them for spreading throughout the world |
|  | United Kingdom (2,81,536) | 4.1362 (11,645) | 95.8637 (2,69,891) | huge difference | great work another home vaccinated carehomevaccination how did it go any lessons for others |
|  | United States (10,57,547) | 6.1715 (65,267) | 93.8284 (9,92,280) | get vaccinated | new zealand has gone days without community transmission of — heres how they did it |
|  | Others (7,19,913) | 4.8426 (34,863) | 95.1573 (6,85,050) | i had the vaccine in china on sunday no problems vaccinated covidvaccine sinovac covid | guarantee a worldwide access to the vaccine treatments and tests to end the pandemic support the trips waiver at the wto ensuring so that any nation can produce or buy sufficient and affordable doses of vaccines treatments and tests peopleoverprofit ruminfin |
| **Waves** | First Wave (1,28,025) | 3.2720 (4,189) | 96.7279 (1,23,836) | truth coronavirus bekind kingston ontario | vp is a deeply immoral man masquerading as a christian |
|  | Delta Wave (5,33,412) | 7.0442 (37,575) | 92.9557 (4,95,837) | covid infections surge despite high vaccination rate vaccines were made first then the virus says rickwiles of true news dr shankara chetty of south africa goal is to vax world to kill billions dr robert malone georgia guidestones | on omicron who shares a new concern ndtv |
|  | Omicron Wave (88,170) | 5.4417 (4,798) | 94.5560 (83,372) | walking the covid trial vaccinationdone✔ vaccin vaccination walk corona coronadebat unsafe scary deaths uneasy | with the midterms approaching wraps up |

| | | | | staysafe stayhealthy | |
| | Others (14,93,905) | 4.9063 (73,296) | 95.0936 (14,20,609) | heading into mbstadium to get a behindthescenes look at the vaccine set up they've created as part of the partnership with fultoninfo to distribute the vaccine details at | great to see palkisu back on screen 💚 she defeated covidprayers and hope for those who still are fighting |

**Table 3:** Descriptive Analysis of Cohort. [Note: Timeline for waves. First Wave: Jan'20 to Jun'20; Delta Wave: Apr'21 to Oct'21; Omicron Wave: Dec'21 to Mar'22]

### Compendium of model performance for misinformation classification

In this paper, we experimented with multiple frameworks for detecting misinformation in tweets. Among the models trained on the discovery cohort, Word2Vec+BiLSTM classified the tweets with accuracy, precision, and recall of 0.92, 0.93, and 0.92 respectively, outperforming all the other models investigated in this work. fastText+BiLSTM performed second best on the discovery cohort with the accuracy, precision, and recall of 0.91, 0.91, and 0.91 respectively. The performance assessment on the discovery cohort for other pipelines has been depicted in Table 4. Owing to the best results on the discovery cohort (Supplementary Fig. 1a), Word2Vec+BiLSTM, fastText+BiLSTM, and GloVe+BiLSTM were chosen for retrospective validation on the randomly sampled retrospective cohort. fastText+BiLSTM gave an accuracy of 87%, followed by GloVe+BiLSTM and Word2Vec+BiLSTM, which resulted in an accuracy of 76% (Supplementary Fig. 1b) on the classification task.

| Model | Accuracy | Precision | F1-Score | Recall |
|---|---|---|---|---|
| Word2Vec + BiLSTM | 0.92 | 0.93 | 0.92 | 0.92 |
| fastText + BiLSTM | 0.91 | 0.91 | 0.91 | 0.91 |
| Glove + BiLSTM | 0.91 | 0.91 | 0.91 | 0.91 |
| BERT +Bi LSTM | 0.87 | 0.86 | 0.86 | 0.87 |
| SapBERT from PubMedBERT with tokenizer | 0.81 | 0.84 | 0.82 | 0.81 |
| spanbert-large-cased | 0.79 | 0.79 | 0.79 | 0.79 |
| roberta-base | 0.79 | 0.74 | 0.76 | 0.79 |
| biosyn-sabert-bc5cdr-disease | 0.76 | 0.83 | 0.78 | 0.76 |
| xlm-roberta-base | 0.75 | 0.8 | 0.77 | 0.75 |
| emotion-english-distill roberta-base | 0.75 | 0.8 | 0.77 | 0.75 |

**Table 4:** Metric Evaluation of models. Word2Vec + BiLSTM performed the best followed by Glove+BiLSTM and Fast text + BILSTM

## Deciphering emotions among misinformation tweets across top models

We analyzed the top performing models (Word2Vec+BiLSTM, fastText+BiLSTM, and GloVe+BiLSTM) on both cohorts for deciphering human behavior in relation to misinformation tweets. The results on the discovery cohort (Fig. 1a, Fig. 1b) depicted that the Word2Vec+BiLSTM model classified "Trust" with the highest sensitivity and specificity of 0.91 and 0.95 respectively. In contrast, it classified "Sadness" and "Fear" with the lowest sensitivity and specificity of 0.72 and 0.89 respectively. While classifying a tweet of emotion "Trust" as misinformation, the false positive rate was 3.6%, while the false positive rates for "Sadness" and "Fear" were 4.89% and 8.33% respectively. The model fastText+BiLSTM classified the emotion "Surprise" with the highest sensitivity and specificity of 0.88 and 0.96 respectively, while it classified "Anger" and "Fear" with the least sensitivity and specificity of 0.68 and 0.91 respectively. While on the retrospective cohort, we generated labels for misinformation across all the tweets. The analysis (Fig. 1c, Fig. 1d) indicated that fastText+BiLSTM predicts "Disgust" as the most predominant emotion, followed by "Anticipation" and "Joy", while the GloVe+BiLSTM model predicts "Anticipation" as the principal emotion, followed by "Disgust" and "Joy". The confusion matrix for emotion classification generated by Word2Vec+BiLSTM and fastText+BiLSTM has been supplemented in Supplementary Table 5.

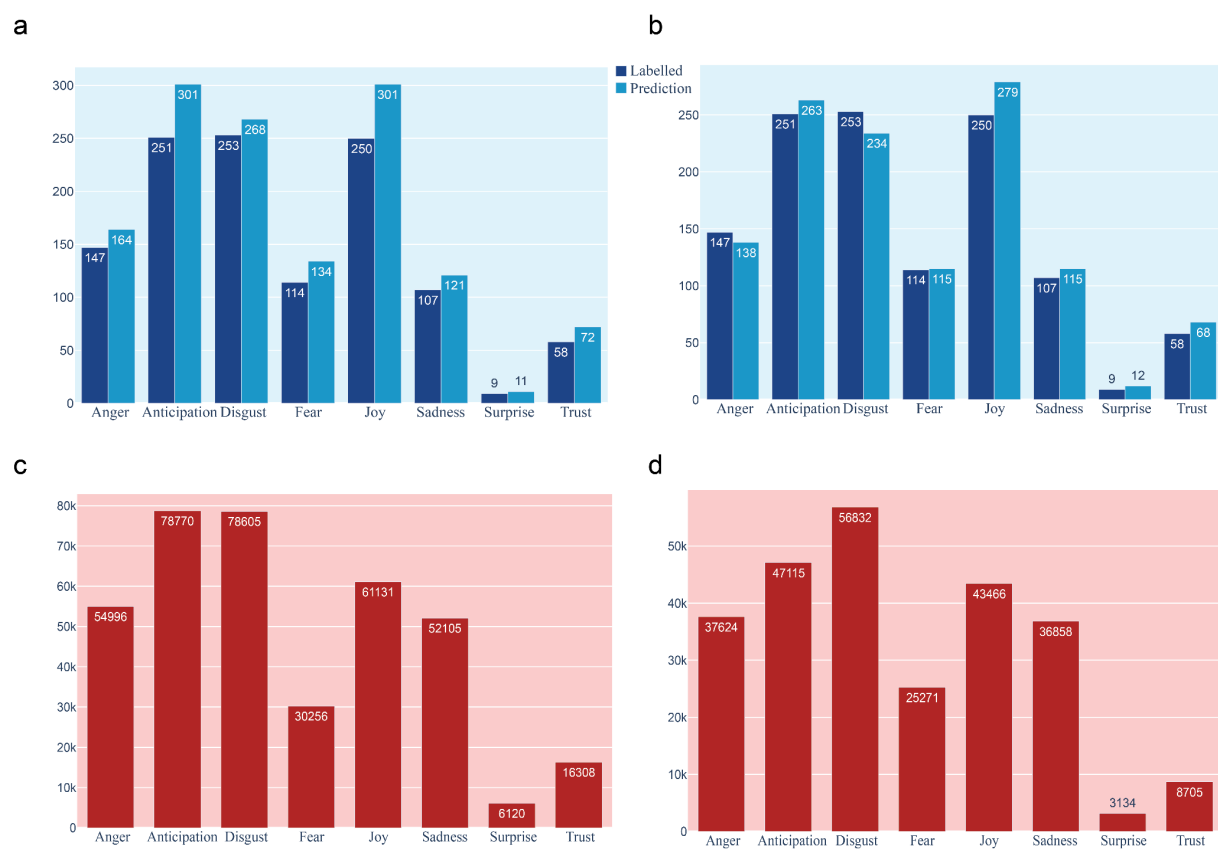

**Fig. 1**: **Distribution of misinformation tweets (labelled and predicted) over 8 emotions across both discovery and retrospective cohorts.** The predominance of emotion in misinformation tweets is predicted by

top-2 models: (a) Word2vec+BiLSTM, (b) fastText+BiLSTM. Profiling based on emotion suggested misinformation tweets with "Surprise" emotion were most accurately predicted by Word2Vec+BiLSTM with an error rate of 0.316%, whereas "Joy" emotion was most incorrectly predicted by the Word2Vec+BiLSTM model with an error rate of 9.222 %. The predominance of emotion in misinformation tweets predicted by top-2 models, (c) fastText+BiLSTM and (d) GloVe+BiLSTM on the retrospective cohort. We have generated predicted labels from these top models on ~2.3M tweets. Both models predicted "Disgust" and "Anticipation" as the most prominent emotions in misinformation tweets.

## Stratification of misinformation tweets based on Spatio-temporal analysis

In this analysis, the stratification of misinformation was done based on spatial and temporal events to determine its spread across the globe and identify predominant emotions that acted as the carriers of misinformation. In the discovery cohort, it was observed that 17.52% of tweets originating from the United States of America were categorized as misinformation. Canada had 15.16% of tweets classified as misinformation, followed by the United Kingdom with 14.65% and India with 9.66%. (Fig. 2a). In the retrospective cohort, 6.16% of tweets from the United States of America were classified as misinformation, with the United Kingdom having the second-highest percentage at 4.14%. India and Canada followed with 4.38% and 4.87% respectively (Fig. 2e). A granular analysis was conducted to understand the implication of misinformation tweets in influencing people's behavior across countries that acted as hotspots for spreading misinformation. For the discovery cohort, analysis performed on the true labels (Fig. 2b) depicted that Joy was the predominant emotion for most misinformation-dense countries (3 out of the top 5). For the United States, Canada, and Australia, the emotion "Joy" was found to amplify misinformation, leading to misinformation rates of 49.07%, 56.14%, and 54.05%, respectively, for tweets associated with the emotion "Joy". The analysis of the predicted labels from Word2Vec+BiLSTM (Fig. 2c) and fastText+BiLSTM (Fig. 2d) models for the discovery cohort suggested that Disgust was predominant in 4 out of 5 misinformation dense countries, followed by Anticipation. The analysis of the annotations of the retrospective cohort done by fastText+BiLSTM (Fig. 2f, Fig. 2g) shows that Disgust emerged as the most prevalent emotion, resulting in misinformation rates of 55.63%, 54.5%, and 49.16% for tweets from Australia, the United States, and the United Kingdom respectively. Heatmap was generated to demonstrate the percentage of misinformation carried by each of the eight emotions across countries where misinformation was most prevalent. The analysis of the labels generated by the GloVe+BiLSTM (Fig. 2h) model illustrated that Disgust was the predominant emotion in 3 out of 5 misinformation-dense countries, followed by Anticipation in 3 out of 5 countries, which have the highest number of misinformation tweets. The results depicted how different countries contributed to global misinformation. Also, the predominant emotions changed concerning locations, which deciphered that there were different carriers of misinformation across geography.

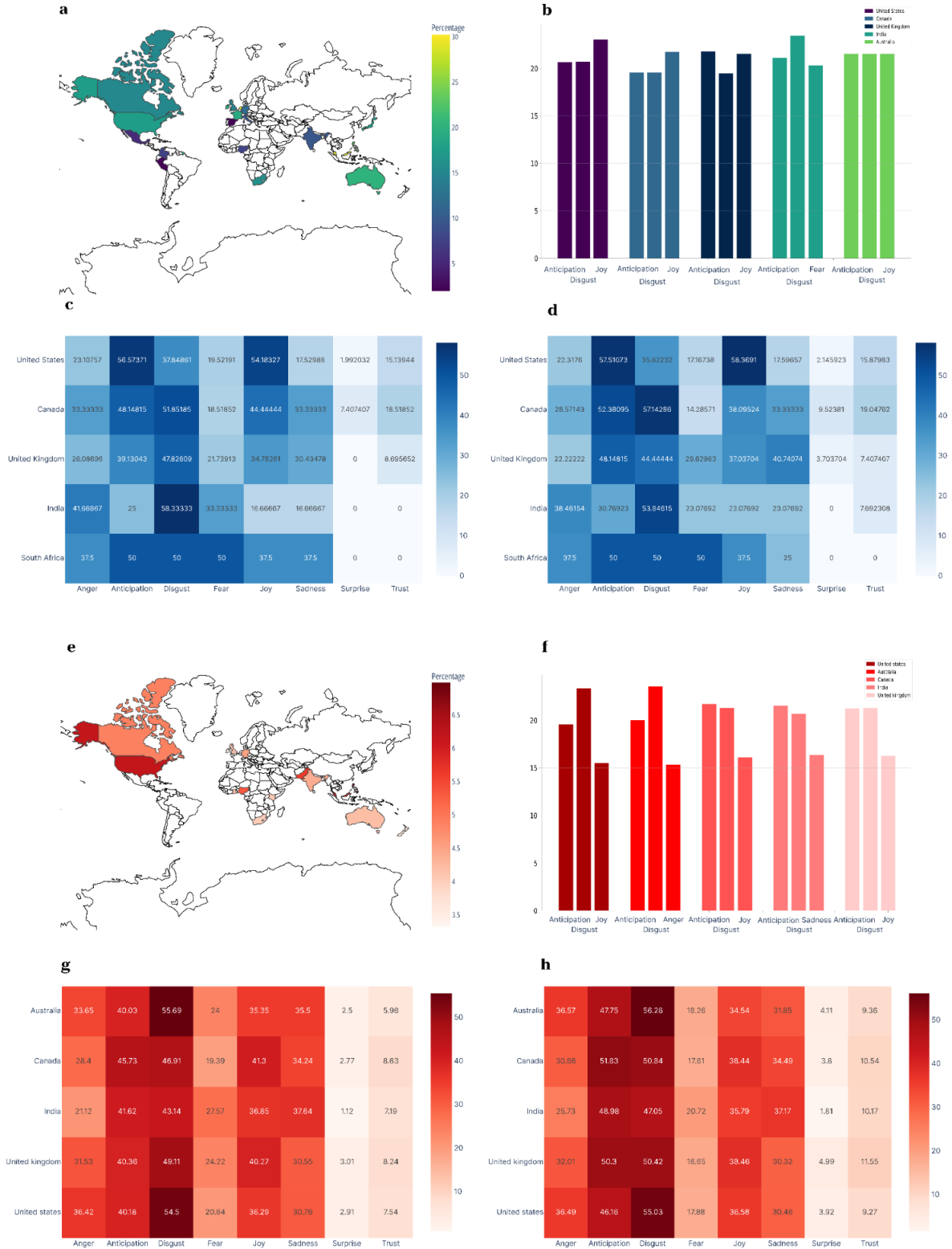

Fig. 2: Spatio-temporal analysis of misinformation tweets. (a) Stratification of misinformation tweets for COVID-19 worldwide (considering countries with more than 50 total tweets available). (b) Deciphering the emotion for misinformation tweets across top-5 countries where misinformation tweets were most dominant. We observed that for all top-5 countries, the predominance of disgust, joy, and anticipation emotions was found, whereas, for India, the top-3 emotions were fear, anticipation, and disgust. We have demonstrated the percentage

of the top 8 emotions across top-5 countries in the figure below. (c) A heatmap of emotions across top-5 countries shows that "Anticipation" was the predominant emotion in the USA for misinformation tweets classified by the Word2Vec+BiLSTM model. In contrast, "Disgust" was the predominant emotion in the UK and India. (d) A heatmap of emotions across top-5 countries shows that "Joy" was the predominant emotion in the USA for misinformation tweets classified by the fastText+BiLSTM model. In contrast, for the UK and India, "Anticipation" and "Disgust" were predominant emotions respectively. (e) Stratification of misinformation tweets predicted by fastText+BiLSTM in the retrospective cohort for COVID-19 worldwide (considering countries with more than 8000 tweets available to maintain a ratio with the discovery cohort). (f) Deciphering the emotion for misinformation tweets predicted by fastText+BiLSTM across top-5 countries where misinformation tweets were most dominant. We observed that for all top-5 countries, the predominance of "Disgust" and "Anticipation" emotions was found. For India, the top-3 emotions were "Anticipation", "Disgust" and "Sadness". We have also demonstrated the percentage of the top 8 emotions across top-5 countries in the figure below. (g) A Heatmap of emotions across top-5 countries shows that "Disgust" was the predominant emotion in all 5 countries for misinformation tweets classified by the fastText+BiLSTM model. (h) A Heatmap of emotions across top-5 countries also depicts that "Disgust" was the predominant emotion in the USA, UK and Australia for misinformation tweets classified by the Glove+BiLSTM model, whereas for India and Canada, "Anticipation" was the predominant emotion.

**Association of misinformation rates and major pandemic events**

This analysis was performed to study the misinformation rate against major pandemic events during COVID-19, such as vaccination and the number of positive cases. With an upsurge in vaccination rate during the timeline from January 2021 to July 2021, a peak in misinformation was also observed across both the cohorts for India and the United States. While bursts of misinformation were observed in July 2021 for the United States across the discovery cohort (Supplementary Fig. 2b), the results on the retrospective cohort depicted a smoother relationship between misinformation and vaccination (Fig. 3a), with the number of misinformation tweets in the United States ascending within three months after the uprise of vaccination rate. The misinformation uprise was seen to lag the vaccination rate in the United States, with the increase in misinformation tweets in the period of July 2021 to January 2022 contributing to 22% of total misinformation tweets in the United States. It was also observed that the misinformation rate increased in proportion to the case rate in the United States throughout the timeline of the retrospective cohort (Fig. 3b). In contrast to the results obtained for the United States, misinformation was observed to have a lead relationship with the vaccination rate for India across both cohorts (Fig. 3c, Supplementary Fig. 2d). The analysis of the retrospective cohort depicted an increase in misinformation tweets from January 2021 to July 2021, constituting 58% of total misinformation tweets in India, which was followed by a rise in vaccination rate from July 2021 to October 2021. The total number of positive cases for India showed that it had a direct correlation with the increase in misinformation (Fig. 3d). The total positive cases and the number of misinformation tweets have an upsurge from January 2021 to July 2021, with the number of misinformation tweets in this interval contributing to 58% of total misinformation tweets for India. The time series for misinformation and case rate for the United States (Supplementary Fig. 2a) and India (Supplementary Fig. 2c) have been supplemented. The fine-granularity time series analysis of vaccination and misinformation riding on predominant emotions in the United States and India has been discussed in the next section. Further, category classification was performed to understand the topics in misinformation tweets at different periods.

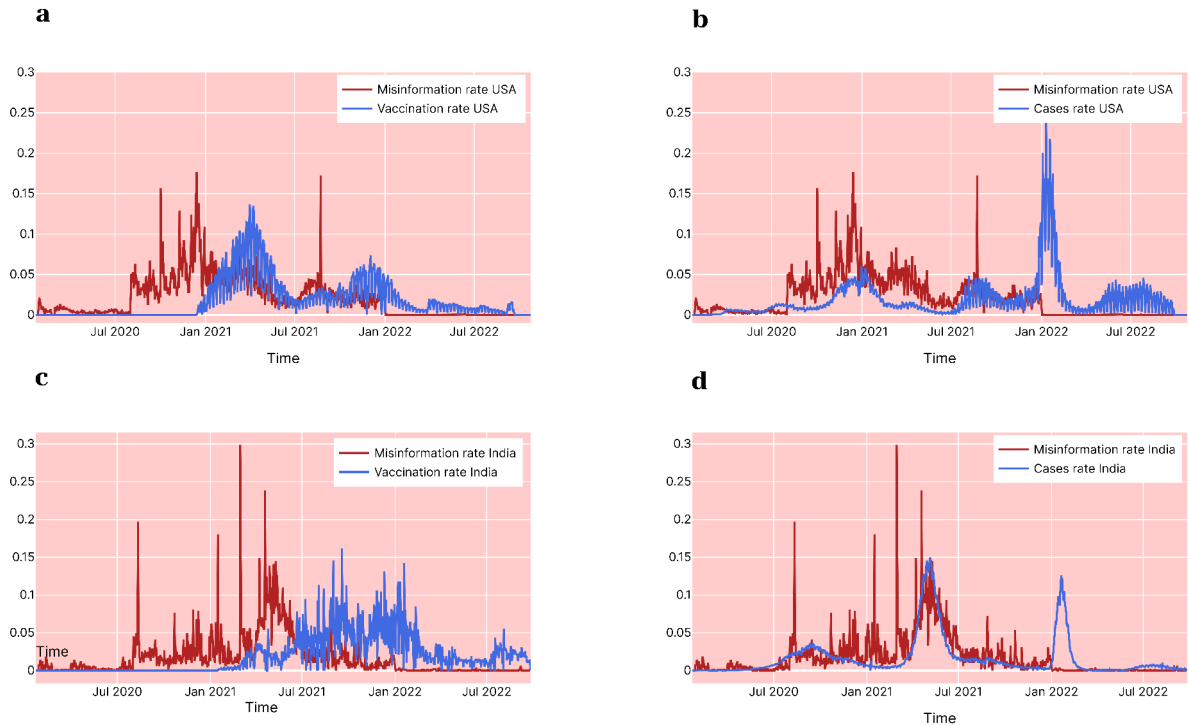

**Fig. 3: Time series plots of misinformation rates against pandemic events for the retrospective cohort.** (a) Time series plot of misinformation rate with vaccination rate for the United States inferred from Retrospective data. (b) Time series plot of misinformation rate with the number of cases per day for the United States inferred from Retrospective data. (c) Time series plot of misinformation rate with vaccination rate for India inferred from Retrospective data. (d) Time series plot of misinformation rate with the number of cases per day for India inferred from Retrospective data.

### Effect of Differential Carriers of Misinformation

Following the temporal analysis of misinformation with vaccination rates across India and the United States, we examined the relationship between vaccination rate and misinformation tweets that carried dominant emotions. The analysis aimed to infer the effects of specific predominant emotions on the vaccination rate across time.

The analysis for the United States on the retrospective cohort illustrated that there were different carriers of misinformation over time that had different effects on the vaccination rate. 13195 misinformation tweets were observed in the United States from July 2021 to December 2021. The top three emotions, Disgust, Anticipation, and Joy, contributed to 59%, 38%, and 33% of the total misinformation tweets for this period respectively. Misinformation tweets riding on disgust (Fig. 4a) and anticipation (Fig. 4b) were seen to lag the vaccination rate. In contrast, misinformation tweets with joy as the primary emotion was observed to rise at nearly the same rate as vaccination (Fig. 4c).

The three predominant emotions observed in India across the retrospective cohort included disgust, anticipation, and sadness. Time series analysis of misinformation riding on each emotion (Fig. 4d, Fig. 4e, Fig. 4f) depicted that misinformation was leading the vaccination rate, with tweets increasing from January 2021 to June 2021, right before the ascent of the vaccination rate in India. Among the 4589 misinformation tweets analyzed during this period, 41% exhibited anticipation as their predominant emotion, 44% were characterized by disgust as their predominant emotion and 39% were associated with sadness as their predominant emotion. The time series analysis for misinformation with cases and vaccination rate on the discovery cohort for the United States and India is shown in Supplementary Fig. 3.

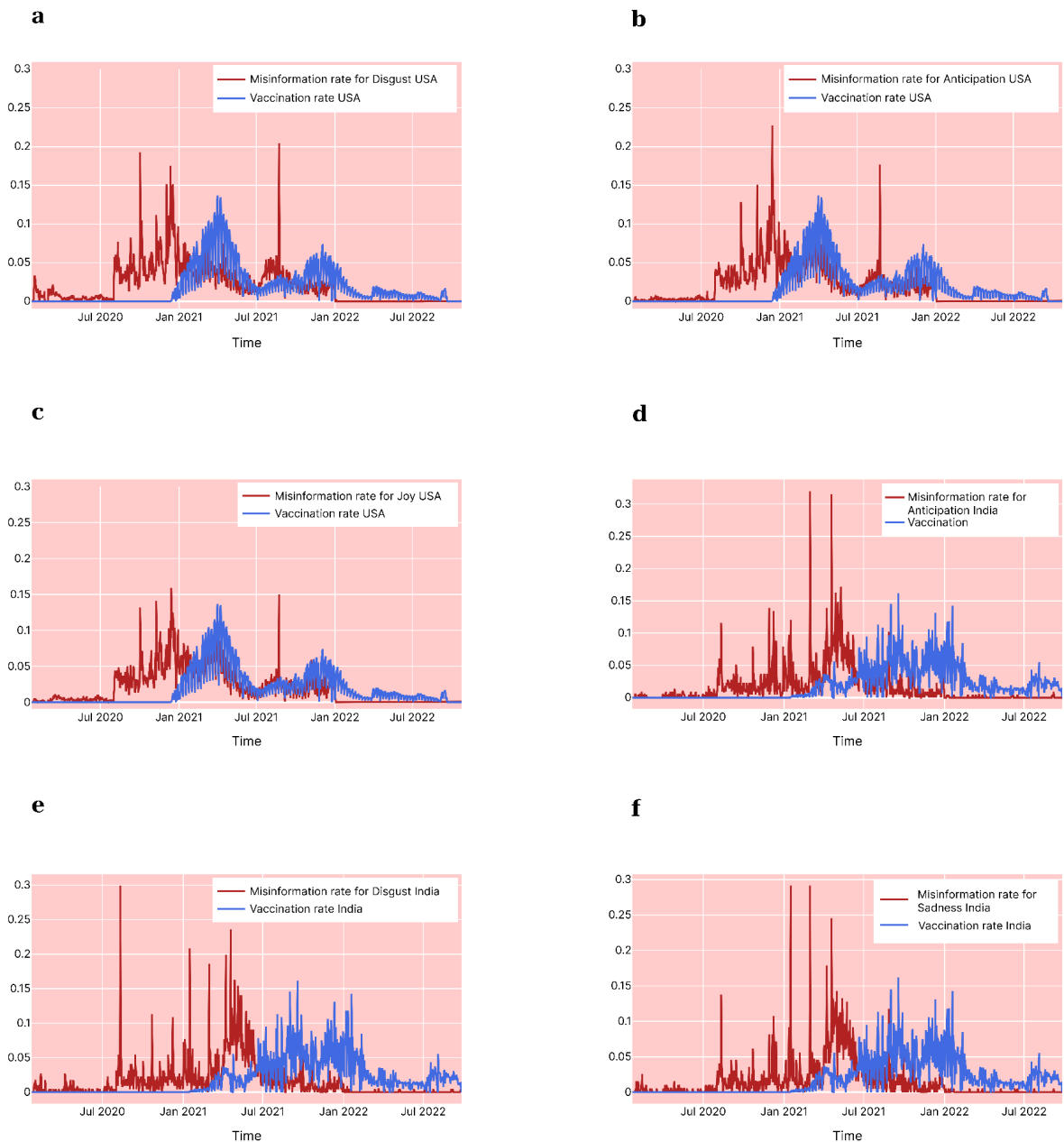

**Fig. 4: Time series plots of top-3 emotions against vaccination rates.** Time series plot of Disgust (a), Anticipation (b), and Joy (c) emotion-misinformation tweets with vaccination rate for the United States respectively inferred from retrospective data. Time series plot of Anticipation (d), Disgust (e), Sadness (f) emotion-misinformation tweets with vaccination rate for India respectively inferred from retrospective data.

## Temporal Evolution of Categories in Misinformation Tweets

In order to analyse the content of misinformation across different timelines of the pandemic, category classification was performed to unfurl the context of tweets that varied in proportion to major pandemic events (like vaccination and cases). Five categories were considered for the study on the retrospective cohort which included "Authorities", "Causes", "Cures", "Symptoms" and "Vaccine". The results show that in India, misinformation related to vaccinations spread more rapidly than any other category with 52.3% of misinformation tweets belonging to the category "Vaccine". However, for the USA, "Symptoms" was the category across which misinformation propagated most swiftly with 47.6% of total misinformation tweets related to this category. "Symptoms" was followed by misinformation tweets in "Vaccines", comprising 28% of total misinformation tweets. It was also observed that the misinformation tweets related to "Vaccine" elevated as the pandemic progressed in India while those for "Symptoms" reduced. A similar trend was evident in the USA with misinformation related to "Vaccine" increasing and that related to "Symptoms" reducing over time. Table 5 shows the result of category classification for misinformation across India and the USA.

| Timeline | Country | Authorities | Causes | Cures | Symptoms | Vaccine | Total |
|---|---|---|---|---|---|---|---|
| Jan 2020 - June 2020 | India | 11(4.38%) | 29(11.55%) | 8(3.19%) | 113(45.02%) | 90(35.86%) | 251 |
| July 2020 - Dec 2020 | India | 62(3.47%) | 222(12.44%) | 54(3.03%) | 714(40%) | 733(41.06%) | 1785 |
| Jan 2021 - June 2021 | India | 62(1.32%) | 320(6.8%) | 62(1.32%) | 1676(35.61%) | 2586(54.95%) | 4706 |
| July 2021 - Dec 2021 | India | 17(1.29%) | 68(5.16%) | 24(1.82%) | 399(30.30%) | 809(61.43%) | 1317 |
| Jan 2020 - June 2020 | USA | 42(2.93%) | 248(17.32%) | 59(4.12%) | 797(55.66%) | 286(19.97%) | 1432 |
| July 2020 - Dec 2020 | USA | 557(3.98%) | 3340(23.89%) | 802(5.74%) | 6311(45.14%) | 2972(21.26%) | 13982 |
| Jan 2021 - June 2021 | USA | 249(2.1%) | 1447(12.21%) | 228(1.92%) | 5904(49.8%) | 4027(33.97%) | 11855 |
| July 2021 - Dec 2021 | USA | 140(1.83%) | 1167(15.26%) | 194(1.54%) | 3623(47.39%) | 2521(32.98%) | 7645 |

**Table 5:** Category Classification for misinformation tweets across India and USA

## Discussion

The COVID-19 pandemic led to significant changes in lifestyle worldwide, resulting in a substantial increase in information sharing on social media platforms such as Twitter, Facebook, etc. The rate of misinformation proportionately increased with the proliferation of information, necessitating the development of systems aimed at detecting misinformation and mitigating its effect on human behavior. This study aimed to characterize emotions as differential carriers of misinformation in the United States and India and their effect on major pandemic events. We have suggested an integrated framework comprising misinformation detection, emotion detection and category classification. To ensure the robustness of the framework, state-of-the-art unsupervised embeddings were used in combination with deep learning models. The misinformation model classified misinformation tweets with an F1-score of 0.87 on the retrospective cohort, depicting the robustness of the architecture. Additionally, our findings indicated an association of misinformation tweets with unwarranted mental health behaviors among individuals, potentially influenced by the pandemic and amplified by the widespread dissemination of false information. Also, the effect of misinformation on major pandemic events was elucidated through category classification. The analysis of categories across misinformation tweets clearly depicted a direct association between the content of misinformation and the pandemic event (vaccination or cases). From July 2020 to January 2021, there was a significant rise in the number of misinformation tweets in India, whose context primarily revolved around the theme of "Vaccine" (41%), which seems to have coincided with the increase in vaccination rates in the country from January 2021. Similarly, in the United States, the predominant category in the misinformation tweets from August 2020 to January 2021 was "Symptoms" (45%). Over this timeline, the misinformation escalation coincided with a rise in the number of positive cases, depicting the theme of misinformation content. Notably, across both India and the United States, misinformation related to "Vaccines" progressed over time while those related to "Symptoms" reduced. This trend could be due to the fact that vaccination campaigns became more widespread, gaining more public attention while people became more aware of the symptoms of the pandemic as it progressed.

In a prior study, Sumit et al. presented misinformation classification on tweets based on deep learning architecture, which included LSTM, BiLSTM, TextCNN, and other transformers-based models. Their results show that LSTM achieved the highest accuracy in classifying misinformation tweets, with an accuracy of 70.7% [26]. We were able to achieve an accuracy of 92% on the discovery cohort and 87% on the retrospective cohort for the misinformation classification task. These results suggested that for Natural Language Understanding (NLU) tasks, utilizing transfer learning for training deep learning models can be advantageous, as it allowed the models to first learn the global context using pre-trained embeddings and then generalize semantic and syntactic features to our dataset, resulting in improved performance. Notably, the hidden state in Bi-LSTM at any given time was not only dependent on the current input and previous hidden state as in vanilla LSTM but also depends on the future input and the future hidden state, giving the models a more comprehensive understanding of the context of the text in our corpus and resulting in better classification decisions. Also, the differences in accuracy outcomes between the three models (Word2Vec+BiLSTM, fastText+BiLSTM, GloVe+BiLSTM) on the discovery cohort and the retrospective cohort can be explained by fastText's ability to incorporate character-level embeddings. This particular feature enabled fastText to better comprehend words that have not been encountered before, resulting in superior performance compared to the other two models when evaluated on the retrospective cohort.

Deciphering the emotional content of tweets is a crucial aspect of analyzing how people were impacted by the COVID-19 pandemic and its related events. By classifying the emotions expressed in tweets, we can gain insights into the behavioral aspects of how people were responding to the pandemic and the various challenges it posed. Sakun Boon-Itt et al. analyzed the public perception of the pandemic on a global scale through sentiment analysis on the Covid-19 tweets collected from December 13, 2019, to March 9, 2020. They found that negative sentiment dominated the discussion about COVID-19 on Twitter with 77.78% of the total tweets classified as negative with "Fear" being the most negative word across tweets and the three themes that were present across the tweets were either the emergency of the pandemic, how to control the pandemic or reports on the pandemic[33]. Their results also showed that public sentiment became more positive as the pandemic progressed. Klaifer Garcia et al. performed the analysis of Covid-19 tweets in English and Portuguese and highlighted the sentiments and the topics that were prevalent in the tweets across Brazil and USA. They inferred that most tweets were associated with the negative sentiment with predominant emotions such as "Anger", "Sadness" and "Fear". They also elucidated that positive sentiments were prevalent at the beginning of the pandemic, possibly because people were optimistic about resolving the pandemic but as time passed, negative emotions emerged and became more predominant[36]. Accordingly, our results indicated that the USA witnessed "Joy", "Disgust" and "Anticipation" as the top 3 emotions across the discovery cohort. For the retrospective cohort, we witnessed the same top three emotions in a different order of prevalence ("Disgust", "Anticipation" and "Joy"). The ambivalent emotions across the discovery cohort can be attributed to the anxiety and unease caused by the increased number of cases, lockdowns, deaths and protests from January 2020 to December 2021[45], stirring negative emotions. Nonetheless, this period also brought about successful clinical trials and approvals of Pfizer vaccines and announcements in the resumption of activities without social distancing [46,47], which could be the reason for the origin of positive sentiments like Joy. The assertion of negative emotions across the retrospective cohort could be because of the surge in hospitalizations based on confounding factors of mutational variants such as the Delta and Omicron[48]. Also, the negativity could be associated with vaccination hesitancy, which was regarded as one of the top health issues[49]. Prasoon et al. presented a study on sentiment analysis of Twitter data related to the lockdown in India during the COVID-19 pandemic. They extracted data from March 25, 2020, to April 14, 2020, and found that 48.69% of the people expressed positive sentiments towards the lockdown, 29.81% had neutral emotions while 21.5% of the people exhibited negative sentiments with regard to the lockdown in India[50]. The analysis demonstrated by D. Chehal et al. inferred "Disgust" had the highest rate of change (178.23%) with a shift in the timeline from second to third lockdown followed by sadness (124.79%)[51]. Our study suggested the same pattern across India with "Anticipation", "Disgust" and "Sadness" misinformation tweets escalating in the retrospective cohort. A possible reason for such sentiments could be the lockdowns and restrictions imposed by the government. Additionally, the fear of contracting the virus might have contributed to these emotional responses.

On analyzing the misinformation tweets, the key takeaways included the findings: (i) Emotion being carrier of misinformation; the emotion classification performed on the retrospective cohort showed that "Disgust" was the most commonly observed emotion across misinformation tweets, comprising 22% (58568) of the total misinformation tweets, thereby acting as the primary carrier of misinformation. It was closely followed by "Anticipation" and "Joy," which comprised 18% (48553) and 17% (44803) of the misinformation tweets in the retrospective cohort respectively, (ii) Different carriers of misinformation across different geographies; analysis across both cohorts, concerning the propagation of misinformation, in different parts of the world showed that different countries had unique sets of emotional triggers that were prevalent in misinformation tweets. While "Disgust" was the most prevalent emotion in misinformation tweets in the United States and Australia, Canada, India, and the United Kingdom observed Anticipation being predominant in their misinformation tweets, (iii) Different carriers of misinformation across time; the emotions expressed in COVID misinformation tweets changed as the pandemic progressed. The misinformation tweets collected from January 2020 to December 2021 across the discovery cohort showed that "Joy" was the predominant emotion in the United States, Canada, and Australia. "Anticipation" and "Disgust" were the predominant emotions in misinformation tweets across the United Kingdom and India. Alternatively, analysis of the retrospective data, which was collected from January 2020 to September 2022 depicted a shift in emotion from "Joy" to "Disgust" in misinformation tweets in the United States and Australia. Misinformation tweets across Canada, India, and the United Kingdom had "Anticipation" as the predominant emotion, (iv) Different carriers of misinformation affected the real world events in a different ways; analysis was performed to illustrate the relationship between predominant emotions and major pandemic events (vaccination and positive cases) across varied geography. It demonstrated that each carrier of misinformation behaved differently with these events. For instance, misinformation riding on disgust increased significantly with the increase in vaccinations in the United States. The misinformation tweets riding on Anticipation and Joy elevated less with the vaccination increase. In India, an immense rise in misinformation tweets riding on "Anticipation" and "Sadness" was observed with an increase in the vaccination rate. In contrast, for the misinformation tweets having "Disgust" as the predominant emotion, it took only a small increase to observe the increase in vaccination rate.

**Limitations of the study**

There are a few limitations to our study. The discovery cohort used for training the deep neural networks was small, comprising only 16567 samples. Although the data was skewed with a number of misinformation tweets being 2838, we have taken measures to ensure the consistency of our findings. We addressed this limitation by validating the misinformation model on the sample of retrospective data that was previously unseen by the models during their training (data from 2022), thereby ensuring the model's robustness. The validation results showed that fastText+BiLSTM gave an accuracy of 87% on the unseen data and hence was able to generalise the results despite being trained on a small dataset. Additionally, we have not made efforts to retrieve the discarded tweets based on Twitter policy. Collocation of the available tweets resulted in a cohort comprising 2.3 million tweets in the retrospective cohort on which analysis was conducted. Hence, the necessity to retrieve missing data was deemed futile for our study. While Plutchik Transformer has been used for emotion classification because of its ability to classify a wide range of emotions and provide detailed classification results, its average F1 score on two benchmark datasets (Company and SemEval) was 0.443 and 0.606 respectively[52]. Furthermore, the Plutchik transformer does not take into consideration secondary emotions such as apprehension, annoyance etc.

## Conclusion

To the best of our knowledge, we present an end-to-end framework for detecting misinformation on social media, while stratifying the behavioral patterns based on geographical and temporal evolution. Our findings are supported by comprehensive analysis derived from ~2.3 million tweets collated during the COVID-19 pandemic. The study's findings provide fresh perspectives on the patterns and motifs associated with COVID-19 misinformation tweets and the behavioral traits of the recipients of this mis(information). The study highlights the importance of focused interventions and educational initiatives to prevent the spread of false information on social media. This framework can also be extended to other threatful diseases such as tuberculosis, mental health diseases or other pandemics and epidemics. The social dynamics (social media engagement) are unlikely to change irrespective of the event. The only variables subjected to change are the emotions that are brought across by the event. Hence, the framework we have suggested in this work will help in generalising the behavioural changes across all events, considering temporal and geographical patterns. To further build upon the findings of this study, a potential future research direction would involve investigating the impact of targeted interventions in mitigating the dissemination of COVID-19 misinformation on social media. Additionally, exploring the factors that influence user engagement with COVID-19 misinformation tweets could provide valuable insights. The findings of this study clearly demonstrate the importance of tackling the issue of COVID-19 misinformation on social media and creating effective solutions for researchers, epidemiologists and policy-makers.

# Supplementary Information

**Train-test split**

Considering data imbalance as a critical factor to be minimized to ensure fair metrics, the discovery cohort was manually split into train and test sets, maintaining the ratio of misinformation to real tweets. The ratio of each source of tweets in the train and test was preserved. The train-test split was performed in the ratio of 80:20.

- CoAID – The tweets in this dataset were annotated and fact-checked from various reliable sources like Lead Stories and PolitiFact. For our study, we have extracted 12842 tweets from this dataset out of which 11414 were labelled as not misinformation and 1428 were misinformation.
- CMU – This dataset comprised tweets with annotations and multiple classes like conspiracy, true treatment and politics. 739 instances were extracted from the CMU dataset. We curated relevant classes into a binary classification based on manual validation. A total of 171 tweets were obtained after the transformation of classes, out of which 35 were not misinformation and 136 were misinformation.
- Antivax - More than 15,000 tweets were annotated as misinformation or general vaccine tweets using reliable sources and were validated by medical experts. A total of 2564 tweets were collected from this dataset out of which 1095 were real and 1469 were misinformation.

**Topic classification**

Latent Dirichlet Allocation (LDA) is an unsupervised clustering method that uses the Dirichlet distribution, which is frequently used in text analysis to find hidden qualities that cannot be examined directly. Words are represented as subjects in this topic modelling approach, and texts are represented as collections of these word topics. LDA will organise the text in the space by their subject. This approach, when applied to preprocessed input text in our study, returned the most relevant phrases for the top n subjects.

| Attributes | Description |
|---|---|
| Id_str | Tweet ID |
| text | Tweet text |
| location | Location of the user |
| created_at | Time the tweet was posted |
| processed_text | Pre-processed tweet text |
| day | Day the tweet was posted |
| month | Month the tweet was posted |
| year | Year the tweet was posted |
| label | Binary label indicating whether the tweet is misinformation or not |

**Supplementary Table 1:** Attributes in the discovery cohort

**Training set**

| Dataset Name | Misinformation tweet | Real tweet |
|---|---|---|
| CMU | 28 | 108 |
| AntiVAX | 1175 | 876 |
| CoAID | 946 | 9131 |
| Human-Annotated | 135 | 853 |

**Supplementary Table 2:** Distribution of (mis)information tweets across training set

**Testing set**

| Dataset Name | Misinformation tweet | Real tweet |
|---|---|---|
| CMU | 7 | 28 |
| AntiVAX | 294 | 219 |
| CoAID | 237 | 2283 |
| Human-Annotated | 16 | 231 |

**Supplementary Table 3:** Distribution of (mis)information tweets across testing set

**Train-test combined**

| Dataset Name | Total Tweets | Misinformation tweet | Real tweet |
|---|---|---|---|
| Test | 3314 | 553 | 2761 |
| Train | 13206 | 2269 | 10937 |

**Supplementary Table 4:** Train-test split of the discovery cohort

| | Word2Vec + BiLSTM | | | | fastText +BiLSTM | | | |
|---|---|---|---|---|---|---|---|---|
| Emotion | True Positive | True Negative | False Positive | False Negative | True Positive | True Negative | False Positive | False Negative |
| Anger | 113 | 635 | 51 | 34 | 101 | 649 | 37 | 46 |
| Anticipation | 250 | 1725 | 66 | 52 | 236 | 1721 | 70 | 66 |
| Disgust | 190 | 1107 | 78 | 63 | 177 | 1128 | 57 | 76 |
| Fear | 89 | 381 | 45 | 25 | 79 | 390 | 36 | 35 |
| Joy | 224 | 1121 | 77 | 26 | 217 | 1125 | 73 | 33 |
| Sadness | 78 | 728 | 43 | 29 | 75 | 731 | 40 | 32 |
| Surprise | 9 | 47 | 3 | 1 | 8 | 46 | 4 | 1 |
| Trust | 53 | 442 | 19 | 5 | 51 | 444 | 17 | 7 |

**Supplementary Table 5:** Confusion matrix for emotion classification by **top-2** models - **Word2Vec+BiLSTM** and **fastText+BiLSTM**. The emotion **"Trust"** is classified with the highest sensitivity of **0.91** and the highest specificity of **0.95** while **"Sadness"** is classified with the lowest sensitivity of **0.72** and **"Fear"** is classified with lowest specificity of **0.89** by the **Word2Vec+BiLSTM** model. The model **fastText+BiLSTM** classified the emotion **"Surprise"** with highest sensitivity and specificity of **0.88** and **0.96** respectively. It classified **"Anger"** with the least sensitivity of **0.68** and **"Fear"** with least specificity of **0.91**.

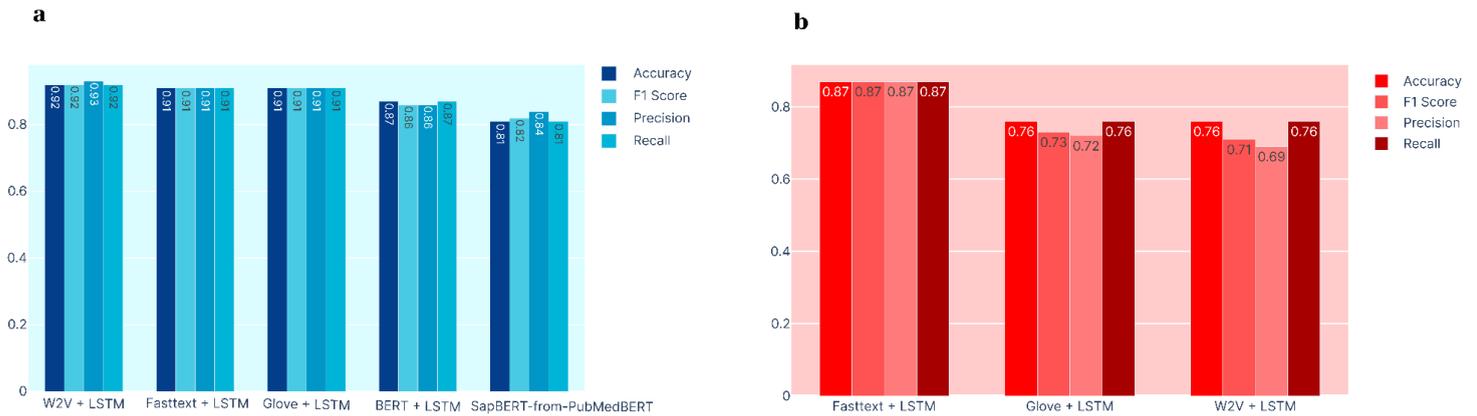

**Supplementary Fig. 1:** (a) Metric Evaluation of the Top five models on our test cohort. **Word2Vec+BiLSTM** performed the best among all other misinformation classifiers models. (b) Retrospective Evaluation of Top three models. **fastText+BiLSTM** performed the best among the top three misinformation classifiers selected on the basis of the test cohort.

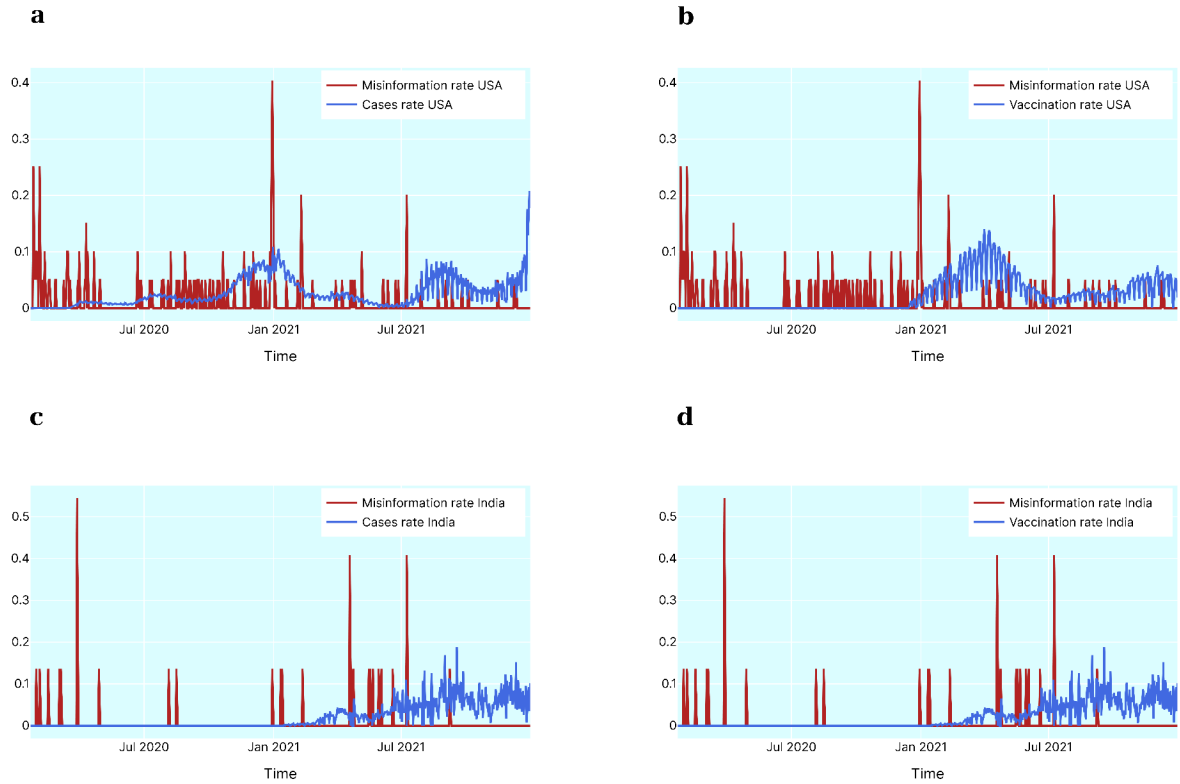

**Supplementary Fig. 2: Time series plots of misinformation rates against pandemic events.** a) Time series plot of misinformation rate with the number of cases per day for the United States. b) Time series plot of misinformation rate with vaccination rate for the United States. c) Time series plot of misinformation rate with the number of cases per day for India. d) Time series plot of misinformation rate with vaccination rate for India. This demonstrates a substantial pattern of misinformation rate with pandemic events (vaccination and number of cases daily). We can infer a leading relationship between Misinformation and Case rates in India.

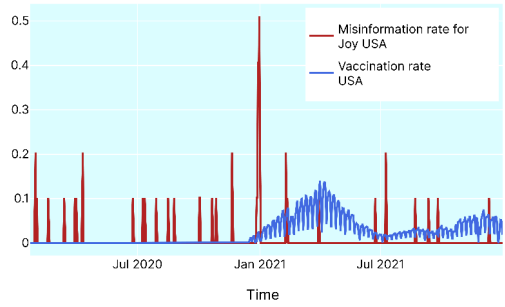
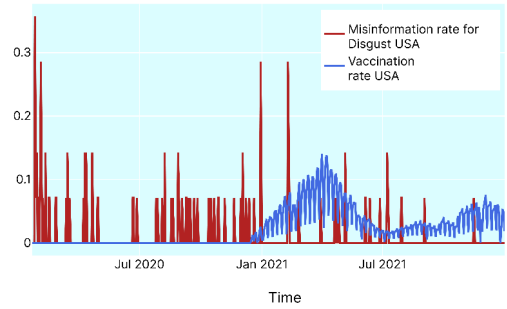
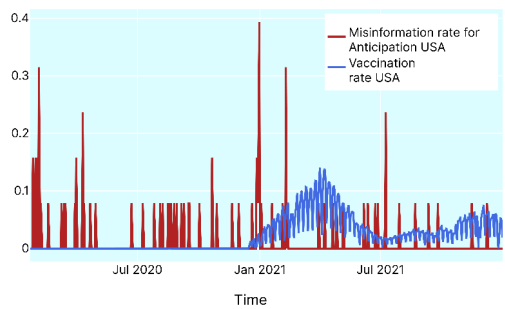
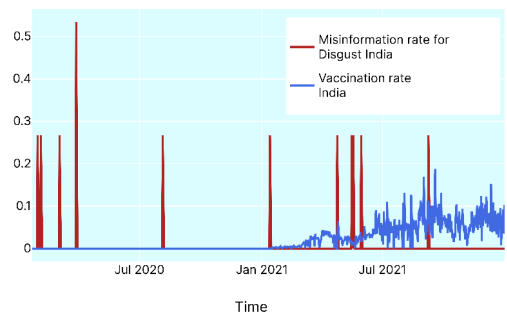
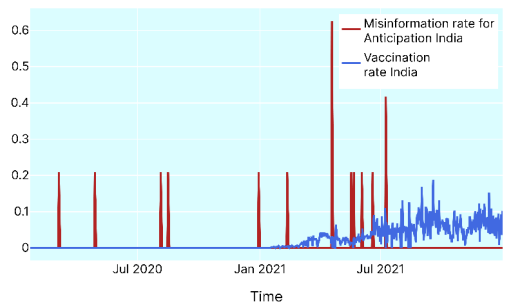
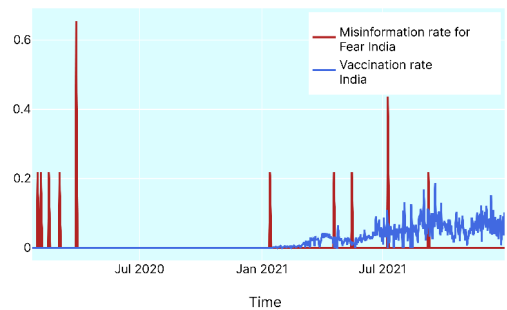

**Supplementary Fig. 3: Time series plots (daily basis) of misinformation tweets of the most occurring emotion against various pandemic events.** Time series plot of Joy (a), Disgust (b), Anticipation (c) emotion misinformation tweets against vaccination rate for the United States respectively. Time series plot of Disgust (d), Anticipation (e), Fear (f) emotion misinformation tweets with vaccination rate for India respectively.